\documentclass[11pt]{article}

\makeatletter
\@ifundefined{papermode}{\def\papermode{preprint}}{}
\makeatother
\usepackage[\papermode]{acl}
\usepackage{times}
\usepackage{latexsym}
\usepackage[T1]{fontenc}
\usepackage[utf8]{inputenc}
\usepackage{microtype}
\usepackage{inconsolata}
\usepackage{graphicx}
\usepackage{booktabs}
\usepackage{array}
\usepackage{multirow}
\usepackage{amsmath}
\usepackage{amssymb}
\usepackage{xcolor}
\usepackage{tikz}
\usetikzlibrary{arrows.meta,positioning,calc,fit,backgrounds}

\definecolor{ours}{HTML}{1A88FF}
\definecolor{oursink}{HTML}{0B5CB8}
\definecolor{rivalgrey}{HTML}{B4B4B4}
\definecolor{paperink}{HTML}{161513}
\definecolor{anuBlue}{HTML}{1A88FF}
\definecolor{anuWash}{HTML}{ECF4FF}
\definecolor{anuChip}{HTML}{F3F4F6}
\definecolor{anuStack}{HTML}{C9D6E6}
\definecolor{anuInk}{HTML}{0A0A0A}
\definecolor{anuMuted}{HTML}{6E6E6E}
\definecolor{anuLine}{HTML}{BDBDBD}


\makeatletter
\newif\ifaclpreprint
\def\acl@preprint{preprint}
\ifx\papermode\acl@preprint \aclpreprinttrue \fi
\makeatother
\newcommand{\sysname}{Anuvani-Diarization}
\newcommand{\pp}{\,pp}
\newcommand{\technicalref}[1]{\ifaclpreprint Section~\ref{#1}\else Appendix~\ref{#1}\fi}
\ifaclpreprint\usepackage{placeins}\fi
\newcommand{\logopath}{figures/logos}
\newcommand{\hfmark}[1][1.2em]{\raisebox{-0.1em}{\includegraphics[height=#1]{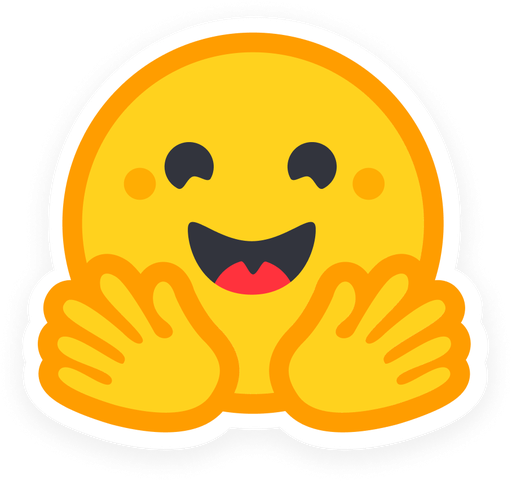}}}

\ifaclpreprint
\makeatletter
\AtBeginDocument{\g@addto@macro\@maketitle{%
  \vskip -0.37in
  \begin{minipage}{\textwidth}
    \centering
    \includegraphics[width=\textwidth]{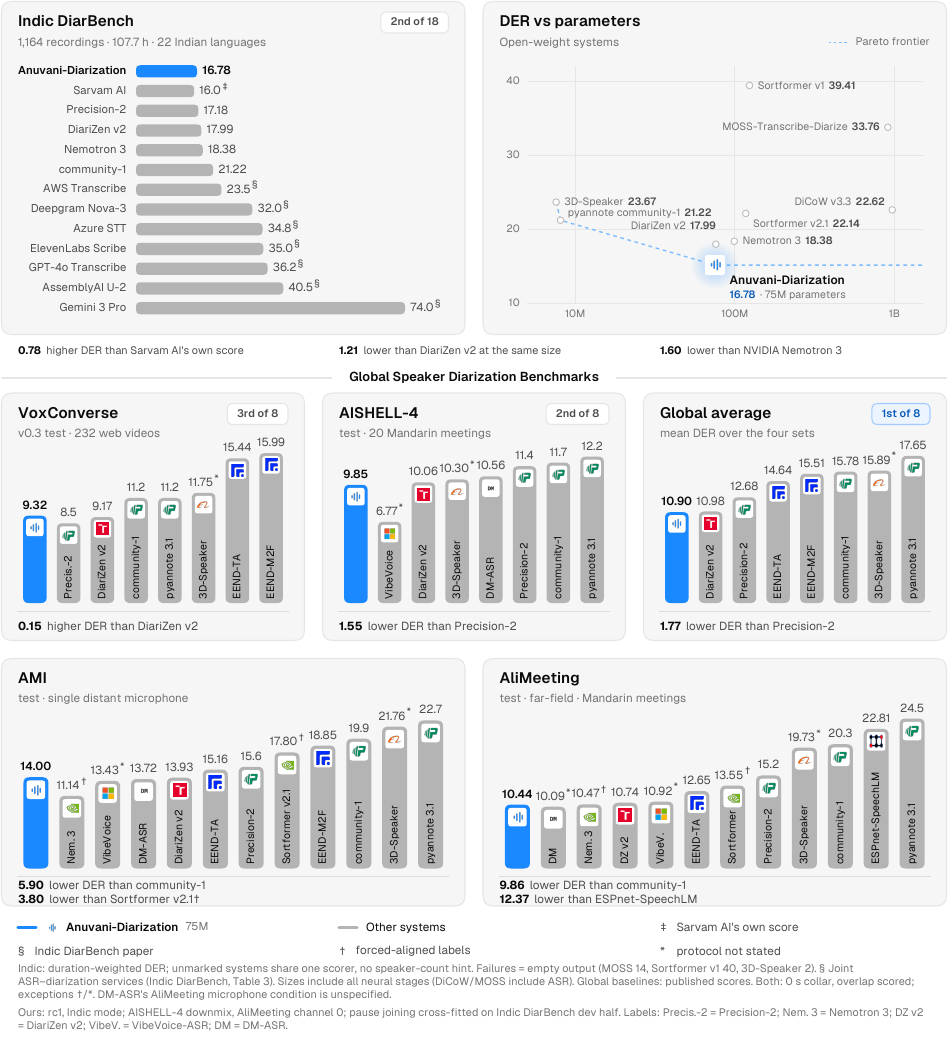}
    \captionof{figure}{Anuvani-Diarization on Indic DiarBench and global benchmarks. DER (\%); bars start at zero.}
    \label{fig:frontpage}
    \par\vspace{0.35em}
    {\small The released model is available at:\par\nobreak
      \vspace{0.2em}\hfmark~\url{https://huggingface.co/CIIL/Anuvani-Diarization}\par}
  \end{minipage}
  \vskip 0.04in}}
\makeatother
\fi

\title{Anuvani-Diarization: Zero Shot Speaker Diarization\\and Annotation Conventions Across 22 Languages}

\ifaclpreprint
\author{\mdseries\itshape Stephen Fernandes, Narayan Choudhary \\[0.15em]
  \upshape Central Institute of Indian Languages \\[0.15em]
  {\normalsize\fontfamily{pcr}\selectfont stephenn.fernandes@gmail.com, n.choudhary@gov.in}}
\else
\author{Anonymous ACL submission}
\fi

\begin{document}
\maketitle
% The taller preprint overview occupies page 1; keep the abstract together on page 2.
\ifaclpreprint\clearpage\fi

% ===================================================================================================
\begin{abstract}
India's 22 scheduled languages are spoken by over a billion people in conversations that mix languages, often within one turn, overlap freely, and take place over telephones, in meetings, and in media. Knowing who spoke when, the task of speaker diarization, is essential to speech applications for such conversation: voice agents conversing with several people at once, speaker attributed transcription of meetings and calls, contact center analytics, and media monitoring. We present a systematic study of speaker diarization across all 22 languages on IndicDiarBench and introduce \sysname{}, which adapts a self-supervised end-to-end segmentation model to Indian conversation with uncertainty aware silver supervision. Evaluated zero shot alongside nine recent diarization systems, it reaches a duration weighted diarization error rate (DER) of 16.78\% on all 1,164 recordings, the lowest of all systems we ran and second only to the best published result (16.0\%), and leads the three strongest rivals in 14 of 22 languages. It matches its parent on six international benchmarks. We also show that the benchmark's utterance level annotation convention is a hidden variable worth 1.9 to 4.1 DER points, depending on the system, and that a convention learned from other Indian transcriptions does not transfer.
\end{abstract}

% ===================================================================================================
\section{Introduction}
\label{sec:intro}

Conversation in India is multilingual by default. The Constitution recognizes 22 languages in its Eighth Schedule \citep{mha2017eighthschedule}, the 2011 Census recorded 19,569 raw mother tongue returns and found that 96.71\% of the population has a scheduled language as its mother tongue \citep{census2011language}, and roughly a quarter of the population reports speaking a second language \citep{sarma2025multilingualism}. Speakers move between languages within a single turn \citep{gumperz1982discourse,bali2014borrowing}, interject, and overlap, and the same conversation may be held over a telephone, a video call or around a table. Every downstream technology that attributes words to people, from meeting transcription to broadcast monitoring and call analytics, first needs to answer the fundamental question of who spoke when.

Until recently there was no benchmark that could measure this capability across India's languages. The DISPLACE challenges introduced diarization of code mixed Indian conversation only for a handful of languages \citep{baghel2023displace,kalluri2024displace}, and IndicDiarBench \citep{mehendale2026indicdiarbench} now provides 108 hours of human corrected, speaker attributed conversation in all 22 scheduled languages, recorded as near field meetings, far field meetings and in-the-wild video. Its first results came from services that transcribe and diarize jointly, the strongest at a diarization error rate (DER) of 16.0\%; dedicated diarization systems, which most research and deployment build on, were not evaluated.

\begin{figure*}[t]
\centering
\includegraphics[width=\textwidth]{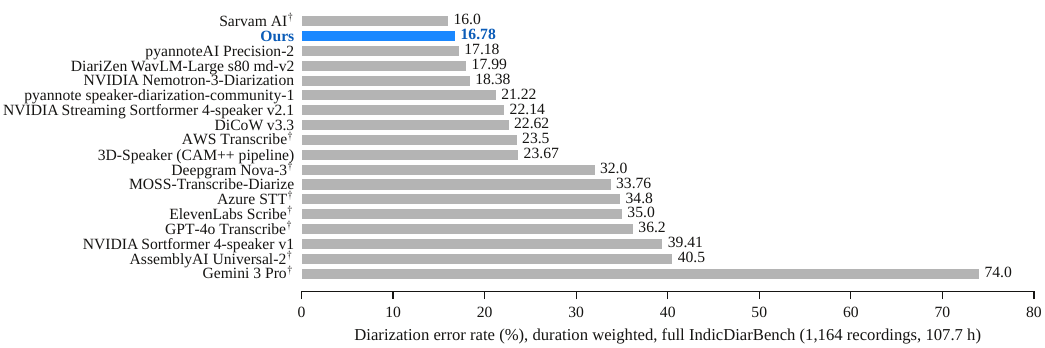}
\caption{\textbf{\sysname{} is the most accurate of all systems we ran on IndicDiarBench.} Duration weighted DER on all 1,164 recordings (zero collar, overlap scored). $^{\dagger}$Joint transcription and diarization services reported by the benchmark's authors \citep{mehendale2026indicdiarbench}; all others were run under our protocol (Table~\ref{tab:main}).}
\label{fig:leaderboard}
\end{figure*}

We ask how the strongest diarization systems perform on Indian conversation when measured blind under one protocol, what, beyond acoustics, determines their errors, and what it would take to lead across the 22 languages. Our contributions are:

\begin{itemize}
\setlength\itemsep{0.15em}
\item \textbf{A blind evaluation of nine diarization systems} on all 1,164 IndicDiarBench recordings, with two independent scorer implementations, bootstrap intervals over source recordings and byte identical sharded inference (\S\ref{sec:protocol}).
\item \textbf{Annotation convention as a hidden variable.} The references segment speech the way transcribers do, into utterances; diarizers segment it at pauses. The gap moves every system by 1.9--4.1\pp{} of DER and the same correction is harmful on word aligned benchmarks, so convention must be an explicit property of a diarizer's output; a convention learned from Indian telephone transcriptions does not transfer to the benchmark (\S\ref{sec:convention}).
\item \textbf{\sysname{}}, an Indic adaptation of a self-supervised end-to-end segmentation model with uncertainty aware silver supervision from channel separated telephone calls (\S\ref{sec:system}). Evaluated zero shot, with no part fitted or selected on the benchmark, it reaches 16.78\% duration weighted DER, the lowest of all systems we ran (Figure~\ref{fig:leaderboard}), and leads the three strongest other systems in 14 of the 22 languages (Table~\ref{tab:languages}).
\item \textbf{Generality}: the adapted network matches its parent on six international benchmarks (\S\ref{sec:beyond}).
\end{itemize}

% ===================================================================================================
\section{Diarizing India: A Linguistic Perspective}
\label{sec:linguistics}

Diarization is usually treated as a language independent task: a speaker's voice is assumed to be the same regardless of the language they speak. India tests this assumption in four ways.

\paragraph{Scale and diversity.}
The 22 scheduled languages range from Hindi, the mother tongue of 528.3 million people (43.63\% of the population), to Sanskrit, returned as a mother tongue by 24,821 people \citep{census2011language}. Eleven of them have more than thirty million mother tongue speakers each. The country's Greenberg diversity index is 0.930: two randomly chosen residents would almost always have different native languages \citep{sarma2025multilingualism}. IndicDiarBench mirrors this distribution only partly. Eight of its languages (Bengali, Gujarati, Hindi, Kannada, Marathi, Punjabi, Tamil and Telugu) each contribute about four hours of near field meetings and appear in far field and in-the-wild recordings, while the fourteen others contribute about 1.5 hours of near field meetings each, and twelve of them appear in no other condition \citep{mehendale2026indicdiarbench}. A diarizer for India must therefore work well on languages for which almost no conversational training data exists.

\paragraph{Four families, one linguistic area.}
The scheduled languages belong to four families: fifteen are Indo-Aryan, four Dravidian (Kannada, Malayalam, Tamil, Telugu), two Tibeto-Burman (Bodo, Manipuri) and one Austroasiatic (Santali); they are written in more than ten scripts (Table~\ref{tab:languages}). Centuries of contact have made South Asia a linguistic area in which unrelated languages share structural and phonetic features \citep{emeneau1956india,masica1976defining}, most audibly the contrast between retroflex and dental consonants and, in Indo-Aryan, a four way laryngeal contrast among stops \citep{ladefoged1996sounds}. Some features are local: Punjabi has lexical tone \citep{hussain2020punjabi}, and Manipuri (Meitei) contrasts high and low lexical tones \citep{chelliah1997meithei}. Pitch and timing, two cues that speaker embedding models exploit, thus carry linguistic as well as personal information. Even the English spoken in these conversations is shaped by the speaker's first language \citep{kachru1983indianization,sirsa2013indianenglish}. For speaker diarization, this means that a speaker model trained mostly on English must separate who is speaking from which language, and which variety of English, is being spoken.

\paragraph{Conversational practice.}
Multilingual speakers switch languages within a given conversation for discourse purposes \citep{gumperz1982discourse}, and Hindi English mixing reaches into almost every register \citep{bali2014borrowing,sitaram2019survey}. IndicDiarBench's annotators transcribe this mixing explicitly, and its authors describe the corpus as rich in ``English code mixing, dialectal variation, and frequent speaker overlap'' \citep{mehendale2026indicdiarbench}. Turn taking is universal in its basic organization but culturally variable when it comes to timing \citep{sacks1974simplest,stivers2009universals,levinson2016turn}: transitions that occur between speakers are short, and a substantial share of them have overlaps rather than gaps \citep{heldner2010pauses}. In IndicDiarBench this share of overlapped speech varies from 6.1\% (Punjabi) to 24.7\% (Maithili) across languages; in-the-wild video has the least overlap (6.5\%), and near field meetings, whose participants were given debate topics, have the most overlap \citep{mehendale2026indicdiarbench}.

\paragraph{The utterance as a unit of annotation.}
Most importantly for this paper, a diarization reference embodies a theory of what a turn is. Word aligned references, typically derived from forced alignment, place segment boundaries at acoustic pauses. IndicDiarBench's references are built by professional transcribers who correct speaker attributed, segment level transcriptions \citep{mehendale2026indicdiarbench}; their unit is the transcribed utterance, and short pauses which are within an utterance are not considered segment boundaries. Section~\ref{sec:convention} shows that this linguistic choice, rather than any acoustic property, is one of the largest single factors in diarization error on the benchmark.

\begin{table*}[t]
\centering
\small
\setlength{\tabcolsep}{4.2pt}
\begin{tabular}{@{}l l l r r c r r r c r@{}}
\toprule
& & & \textbf{L1 speakers} & \textbf{Bench} & & \textbf{Overlap} & \multicolumn{4}{c}{\textbf{DER (\%), pooled}} \\
\cmidrule(l){8-11}
\textbf{Language} & \textbf{Family} & \textbf{Script} & \textbf{(M), 2011} & \textbf{hours} & \textbf{Cond.} & \textbf{(\%)} & \textbf{Ours} & \multicolumn{2}{c}{\textbf{Best other}} & \textbf{$\Delta$} \\
\midrule
Assamese & IA & Assamese & 15.3 & 1.5 & N & 13.9 & \textbf{22.43} & 23.53 & DZ & $-$1.10 \\
Bengali & IA & Bengali & 97.2 & 12.6 & NFW & 7.8 & 17.09 & \textbf{16.76} & P2 & $+$0.34 \\
Bodo & TB & Devanagari & 1.5 & 1.6 & N & 15.2 & 24.34 & \textbf{22.77} & NM & $+$1.56 \\
Dogri & IA & Devanagari & 2.6 & 1.4 & N & 24.2 & \textbf{21.07} & 21.99 & DZ & $-$0.92 \\
Gujarati & IA & Gujarati & 55.5 & 11.1 & NFW & 7.6 & \textbf{13.08} & 13.65 & P2 & $-$0.57 \\
Hindi & IA & Devanagari & 528.3 & 10.7 & NFW & 16.6 & 16.63 & \textbf{16.34} & NM & $+$0.29 \\
Kannada & Dr & Kannada & 43.7 & 8.5 & NFW & 15.2 & \textbf{16.24} & 16.77 & P2 & $-$0.53 \\
Kashmiri & IA & Perso-Arabic & 6.8 & 1.1 & N & 21.2 & \textbf{19.41} & 19.89 & NM & $-$0.48 \\
Konkani & IA & Devanagari & 2.3 & 1.6 & N & 14.0 & \textbf{20.57} & 20.84 & NM & $-$0.27 \\
Maithili & IA & Devanagari & 13.6 & 1.3 & N & 24.7 & 20.71 & \textbf{18.73} & NM & $+$1.98 \\
Malayalam & Dr & Malayalam & 34.8 & 3.7 & NW & 12.9 & 17.79 & \textbf{17.22} & P2 & $+$0.57 \\
Manipuri & TB & Meitei Mayek & 1.8 & 1.5 & N & 20.6 & \textbf{19.82} & 21.75 & NM & $-$1.92 \\
Marathi & IA & Devanagari & 83.0 & 10.4 & NFW & 11.3 & \textbf{14.36} & 14.54 & P2 & $-$0.18 \\
Nepali & IA & Devanagari & 2.9 & 1.3 & N & 22.9 & \textbf{24.96} & 25.65 & P2 & $-$0.69 \\
Odia & IA & Odia & 37.5 & 3.1 & NW & 11.1 & 13.31 & \textbf{12.23} & NM & $+$1.08 \\
Punjabi & IA & Gurmukhi & 33.1 & 10.6 & NFW & 6.1 & 19.78 & \textbf{18.90} & P2 & $+$0.87 \\
Sanskrit & IA & Devanagari & 0.02 & 1.6 & N & 21.4 & \textbf{17.80} & 19.68 & DZ & $-$1.88 \\
Santali & AA & Ol Chiki & 7.4 & 1.6 & N & 6.5 & \textbf{12.66} & 13.98 & P2 & $-$1.33 \\
Sindhi & IA & Perso-Arabic & 2.8 & 1.5 & N & 16.0 & 18.62 & \textbf{16.82} & NM & $+$1.80 \\
Tamil & Dr & Tamil & 69.0 & 10.0 & NFW & 12.4 & \textbf{14.62} & 15.18 & P2 & $-$0.56 \\
Telugu & Dr & Telugu & 81.1 & 9.3 & NFW & 20.4 & \textbf{21.22} & 22.29 & P2 & $-$1.06 \\
Urdu & IA & Perso-Arabic & 50.8 & 1.6 & N & 12.5 & \textbf{16.24} & 17.75 & P2 & $-$1.51 \\
 
\midrule
\textbf{All 22 languages} & & & & 107.7 & & 12.8 & \textbf{17.11} & 17.58 & P2 & $-$0.47 \\
 
\bottomrule
\end{tabular}
\caption{\textbf{The 22 scheduled languages and \sysname{}'s accuracy on each.} Family (IA Indo-Aryan, Dr Dravidian, TB Tibeto-Burman, AA Austroasiatic), benchmark hours, recording conditions (N near field, F far field, W in-the-wild) and overlap are from \citet{mehendale2026indicdiarbench}; mother tongue speakers from the 2011 Census \citep{census2011language}. ``Best other'' is the most accurate of pyannoteAI Precision-2 (P2), DiariZen md-v2 (DZ) and NVIDIA Nemotron-3-Diarization (NM) on that language; the better of ours and the best other is in bold, and $\Delta$ is ours minus the best other, in percentage points, computed from the unrounded values. Bench hours are hours in the benchmark. \sysname{} leads in 14 of the 22 languages.}
\label{tab:languages}
\end{table*}

% ===================================================================================================
\section{Related Work}
\label{sec:related}

\paragraph{Diarization models.}
Modular diarization systems detect speech, embed short segments and cluster the embeddings, with Bayesian HMM clustering of x-vectors (VBx) as a strong back end \citep{landini2022vbx,park2022review}. End-to-end neural diarization (EEND) predicts speaker activities jointly and processes overlap natively \citep{fujita2019eend}; EEND with vector clustering (EEND-VC) combines local end-to-end segmentation with global clustering, so that recordings of any length and speaker count can be processed effectively \citep{kinoshita2021eendvc}. The powerset formulation casts local segmentation as single label classification over subsets of active speakers \citep{plaquet2023powerset} and underlies pyannote \citep{bredin2023pyannote,pyannote2025community1}; DiariZen replaces its front end with a structurally pruned WavLM \citep{chen2022wavlm,han2025diarizen,han2025pruning} and a Conformer \citep{gulati2020conformer}. NVIDIA Sortformer models are trained explicitly with arrival-time-ordered speaker supervision \citep{park2024sortformer,medennikov2025streamingsortformer}, the approach of NVIDIA Nemotron-3-Diarization \citep{nvidia2026nemotron}.

\paragraph{Evaluation conventions.}
DER is sensitive to how references are made. Benchmarks differ in forgiveness collars and in whether overlap is scored; DIHARD established the strict setting of no collar with overlap scored \citep{ryant2021dihard3}, and the underlying scoring tools differ in how they treat overlapping same speaker references \citep{bredin2017metrics,ryant2018dscore}. Most closely related to our analysis, \citet{horiguchi2026pause} decompose DER into a pause attributable component and a core component, and recent work trains models toward tight segment boundaries \citep{tightboundary2026}. We complement this line by measuring the convention of a multilingual benchmark directly, by showing that its effect is large and system wide, and by testing whether a convention learned from other transcriptions transfers to it.

\paragraph{Speech technology for Indian languages.}
Large data collection efforts have produced corpora and benchmarks for Indian languages, including IndicVoices \citep{javed2024indicvoices}, Kathbath and IndicSUPERB \citep{javed2023indicsuperb}, IndicWav2Vec \citep{javed2021indicwav2vec}, SPRING-INX \citep{nithya2023springinx} and the MahaDhwani pre-training corpus \citep{bhogale2025mahadhwani}; code switched recognition has its own challenge series \citep{diwan2021mucs}. For diarization, DISPLACE 2023 provided 32 hours of code mixed conversation and DISPLACE 2024 158 hours, 38 of them labeled, in five Indian languages \citep{baghel2023displace,kalluri2024displace,mehendale2026indicdiarbench}; IndicDiarBench is the first benchmark to cover all 22 scheduled languages \citep{mehendale2026indicdiarbench}.

\paragraph{Concurrent work.} A related concurrent study evaluates spoken language identification across Indian languages, with emphasis on short-input robustness, calibration and language confusion \citep{fernandes2026anuvanilid}. Its prediction target is the language of an audio segment. In contrast, this work predicts speaker activity over time and studies diarization error and annotation conventions. The two studies share their authors and build on publicly released Indian speech corpora; the public SPRING-INX corpus appears in both (as adaptation data here, as a held-out evaluation set there). Their pretrained backbones, benchmarks and experiments are otherwise distinct. The contributions and evaluations concern different speech-processing tasks.

% ===================================================================================================
\section{Benchmark and Evaluation Protocol}
\label{sec:protocol}

\paragraph{Data.}
IndicDiarBench contains 1,164 recordings (107.7 hours) with 2 to 9 speakers each: near field meetings recorded over an online platform with one close microphone per participant (about 53 hours, all 22 languages), far field meetings (27 hours, eight languages) and in-the-wild YouTube conversations (28 hours, ten languages). These meetings involve 485 speakers from 189 districts; references come from a five stage human pipeline \citep{mehendale2026indicdiarbench}. We use the Hugging Face release at revision 92877ba.

\paragraph{Scoring.}
We follow the benchmark: DER with no collar and overlapped speech scored. We score with NIST md-eval semantics (our scorer is identical to dscore \citep{ryant2018dscore} on every system and set we checked); the scorer merges the small number of overlapping same speaker reference segments in the benchmark. \ifaclpreprint
We report both pooled (micro) DER and the duration weighted average used by the benchmark paper. For recording $i$, let $E_i$ be the total missed speech, false alarm and speaker confusion time, $S_i$ the reference speaker time, and $T_i$ the recording duration:
\begin{equation}
\begin{aligned}
\mathrm{DER}_{\mathrm{pooled}} &= \frac{\sum_i E_i}{\sum_i S_i}, \\
\mathrm{DER}_{\mathrm{DW}} &= \frac{\sum_i T_i(E_i/S_i)}{\sum_i T_i}.
\end{aligned}
\end{equation}
We express both ratios as percentages. Pooled DER weights recordings by reference speaker time; duration weighted DER weights them by recording duration. Confidence intervals come from 5,000 bootstrap resamples of the 590 source recordings (seed 3407).
\else
We report both pooled (micro) DER and the duration weighted average used by the benchmark paper. Confidence intervals come from 5,000 bootstrap resamples of the 590 source recordings.
\fi

\paragraph{Competitor systems.}
We ran nine diarization systems on the same single channel audio with each project's recommended configuration: pyannoteAI Precision-2 through its commercial API \citep{pyannoteai2026precision2}, DiariZen WavLM-Large s80 md-v2 \citep{han2025diarizen}, NVIDIA Nemotron-3-Diarization \citep{nvidia2026nemotron}, pyannote speaker-diarization-community-1 \citep{pyannote2025community1}, NVIDIA Streaming Sortformer 4-speaker v2.1 and NVIDIA Sortformer 4-speaker v1 \citep{park2024sortformer,medennikov2025streamingsortformer}, DiCoW v3.3 \citep{dicow2025}, the 3D-Speaker pipeline \citep{chen2024threedspeaker} and MOSS-Transcribe-Diarize \citep{moss2026transcribe}. All outputs were scored together. For the systems that we ran on several GPUs, a single GPU replay of all 1,164 recordings reproduced the sharded outputs byte for byte.

\paragraph{Zero shot protocol.}
Nothing in our system, \sysname{}, was trained, tuned or selected on IndicDiarBench. The segmentation network was adapted on other recordings (\S\ref{sec:adapt}), audio fingerprint screening found no audio shared with the benchmark, the recipe and checkpoint were fixed on an internal panel of non-benchmark recordings, and its clustering uses the parent model's default settings. The system was frozen before it was first scored on the benchmark, although earlier versions of it had been scored there. The competitor systems were run off the shelf, so every system we report was evaluated zero shot.

% ===================================================================================================
\section{Annotation Convention as a Hidden Variable}
\label{sec:convention}

\begin{figure*}[t]
\centering
\includegraphics[width=\textwidth]{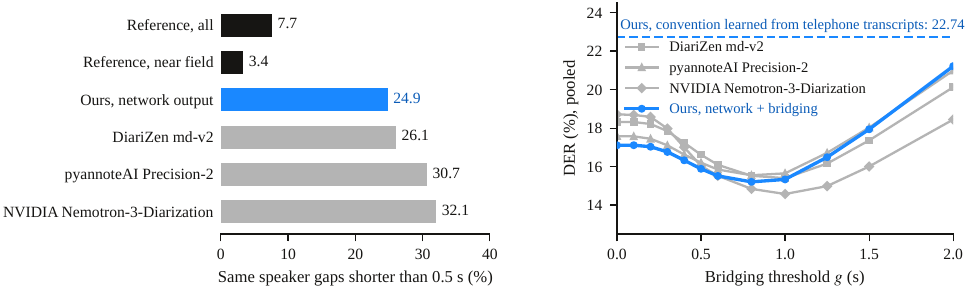}
\caption{\textbf{The benchmark's references segment speech into utterances; diarizers segment it at pauses.} Left: share of same speaker gaps shorter than 0.5\,s on all 1,164 recordings. Right: pooled DER as same speaker gaps shorter than $g$ are bridged ($y$-axis from 12.5\%); dashed: \sysname{} with a convention learned from Indian telephone transcripts, scored once (\S\ref{sec:transfer}).}
\label{fig:convention}
\end{figure*}

\paragraph{Diagnosis.}
Figure~\ref{fig:convention} (left) compares gaps between consecutive same speaker segments. In IndicDiarBench's references only 7.7\% of such gaps are shorter than half a second, and only 3.4\% in the near field meetings. In the output of every diarizer in Figure~\ref{fig:convention} (left), the share is between 24.9\% and 32.1\%. The systems are not wrong about the acoustics: speakers tend to pause inside their turns. They disagree with the references about what counts as a segment, and every such pause inside a reference utterance is scored as missed speech.

\paragraph{Effect.}
Bridging same speaker gaps shorter than a threshold $g$ is a convention change that leaves the acoustic decisions untouched. Its effect is large and consistent (Figure~\ref{fig:convention}, right; Table~\ref{tab:convention}). As a sensitivity analysis, with $g$ chosen on the benchmark itself (a measurement of the convention, not a system), DER falls by 1.90\pp{} for \sysname{}, 2.03\pp{} for pyannoteAI Precision-2, 2.91\pp{} for DiariZen md-v2 and 4.14\pp{} for NVIDIA Nemotron-3-Diarization, the most finely segmenting system we measured, at thresholds of 0.8 to 1.0\,s.

\begin{table}[t]
\centering
\footnotesize
\setlength{\tabcolsep}{2.6pt}
\begin{tabular}{@{}l r r r r@{}}
\toprule
\textbf{System} & \textbf{Raw} & $\boldsymbol{g^{\ast}}$ & \textbf{After} & $\boldsymbol{\Delta}$ \\
\midrule
Ours (\sysname{}) & 17.11 & 0.8 & 15.21 & $-$1.90 \\
pyannoteAI Precision-2 & 17.58 & 0.8 & 15.55 & $-$2.03 \\
DiariZen md-v2 & 18.31 & 1.0 & 15.40 & $-$2.91 \\
NVIDIA Nemotron-3-Diarization & 18.71 & 1.0 & 14.58 & $-$4.14 \\
 
\midrule
\multicolumn{5}{@{}l}{\textit{Word aligned public test sets (ours)}} \\
VoxConverse v0.3 test & 9.32 & 1.0 & 10.00 & $+$0.68 \\
NOTSOFAR-1 eval & 17.41 & 1.0 & 21.63 & $+$4.22 \\
MagicData-RAMC test & 11.13 & 1.0 & 17.14 & $+$6.01 \\
 
\bottomrule
\end{tabular}
\caption{\textbf{Matching the reference convention moves every system, in opposite directions on different benchmarks.} Pooled DER before (Raw) and after bridging same speaker gaps shorter than $g^{\ast}$, the best threshold on the benchmark itself (a sensitivity analysis, not a system result); $\Delta$ is computed from the unrounded values. Bottom: on word aligned references, bridging at $g=1.0$\,s raises DER.}
\label{tab:convention}
\end{table}

\paragraph{A property of the benchmark, not of speech.}
Applied to benchmarks whose references are word aligned, the same bridging raises DER, by 0.68 to 6.01\pp{} on VoxConverse, NOTSOFAR-1 and MagicData-RAMC (Table~\ref{tab:convention}, bottom). There is therefore no convention free output: the best segmentation depends on whose notion of a turn is being scored. Utterance level references suit speaker attributed transcription and word aligned references acoustic analysis, and neither is wrong. We draw three consequences: benchmarks should document their segmentation conventions as precisely as their collar and overlap rules; systems should be compared only under matched output conventions; and a diarizer should expose its convention as an explicit, documented mode rather than silently adopting one.

\ifaclpreprint
\subsection{Learning an annotation convention}
\else
\paragraph{A learned convention does not transfer.}
\fi
\label{sec:transfer}
Can the convention be learned without the benchmark's labels? We fitted a convention decoder\ifaclpreprint\else{} (Appendix~\ref{app:zs})\fi{} on human transcriptions of 302 SPRING-INX telephone calls, which, like the benchmark's references, are speaker attributed utterance spans; chose every setting on 35 held out calls under a protocol fixed in advance; and scored the frozen system once on all 1,164 recordings. DER rose from 17.11\% to 22.74\% (Table~\ref{tab:ablation}): missed speech fell from 8.91\% to 2.89\%, but false alarm rose from 3.55\% to 15.29\%. On the telephone transcripts, bridging kept helping up to the largest gap we allowed (1.5\,s): their transcribers keep an utterance open across longer pauses than the benchmark's annotators do. Two corpora that both segment speech into transcribed utterances thus still disagree about where an utterance ends, so a diarizer can match a benchmark's convention only from references produced the same way.

% ===================================================================================================
\ifaclpreprint
\paragraph{Selection protocol.}
\label{app:zs}
A protocol fixing every candidate and decision rule was committed before any measurement, and a filesystem guard refused every IndicDiarBench file to the selection code. For this experiment only, clustering was chosen on 146 recordings of NOTSOFAR-1, VoxConverse, SPRING-INX, AMI, AISHELL-4 and AliMeeting from 21 VBx configurations, with and without a cannot-link repair that splits clusters whose strongly supported local speakers share a window; the repair at threshold 0.65 and $F_a=0.06$ won (mean pooled DER 12.26\% against 12.51\% for the default). 

\paragraph{Decoder training.}
References for the convention were the human transcript spans of each SPRING-INX call's two channels. The bridging gap was chosen on 35 held out calls from $g\in\{0, 0.25, \dots, 1.5\}$\,s (best: 1.5\,s). The decoder, gradient boosted regression trees \citep{ke2017lightgbm,pedregosa2011sklearn} with 400 iterations, learning rate 0.05, 31 leaves and at least 200 samples per leaf, predicts for every same speaker gap and every segment shorter than 3\,s the change in error time that bridging the gap or dropping the segment would cause, from the local geometry, the recording and the network's own evidence. 

\paragraph{Held-out selection and final evaluation.}
Fitted on the 302 training calls, it beat blanket bridging on the 35 held out calls (duration weighted DER 30.02\% against 32.11\%, paired 95\% CI entirely below zero) and was adopted. The frozen system was scored once on IndicDiarBench: 22.74\% pooled and 22.43\% duration weighted DER.
\fi

\section{\sysname{}}
\label{sec:system}

\ifaclpreprint
% The architecture figure, twin of the Anuvani-LID paper's (figs/arch.tex); preprint-only so the
% anonymous ARR edition keeps the submitted single-column diagram.
\begin{figure*}[t]
  \centering
  \resizebox{\textwidth}{!}{% Anuvani-Diarization segmentation pipeline + adaptation recipe (TikZ; brand palette defined in the
% preamble). Drawn in the exact design of the Anuvani-LID architecture figure, upright as architectures
% are in the field: audio at the bottom, the Indic-adapted segmentation network as a wide framed tower
% (WavLM-Large, pruned to 80 % sparsity, under the Conformer head), the parent model's embedding and
% clustering stages on top, and one Conformer block opened beside the tower. The inset reuses the LID
% block's geometry and residual paths; dimensions follow the released diarization network.
\begin{tikzpicture}[
  font=\sffamily\scriptsize,
  box/.style={draw=anuLine, line width=0.45pt, rounded corners=2.2pt, fill=white, align=center,
              inner sep=2.6pt, minimum height=0.46cm, minimum width=5.2cm},
  hero/.style={draw=anuBlue, line width=0.7pt, rounded corners=2.2pt, fill=anuWash, align=center,
               inner sep=2.6pt, minimum height=0.46cm, minimum width=5.2cm},
  sub/.style={draw=anuLine, line width=0.45pt, rounded corners=2.2pt, fill=white, align=center,
              inner sep=2.2pt, minimum height=0.46cm, minimum width=3.9cm},
  chip/.style={draw=none, rounded corners=2pt, fill=anuChip, align=left, inner sep=2.6pt,
               font=\sffamily\scriptsize, text=anuInk},
  arr/.style={-{Stealth[length=3.2pt,width=2.6pt]}, line width=0.55pt, draw=anuMuted},
  harr/.style={-{Stealth[length=3.2pt,width=2.6pt]}, line width=0.7pt, draw=anuBlue},
  lbl/.style={font=\sffamily\fontsize{7}{8}\selectfont, text=anuMuted, align=center}
]
\def\cx{3.3}\def\W{5.2}
% ---------------- input (bottom up) ----------------
\node[box, minimum height=0.56cm] (wav) at (\cx,0.28) {};
\foreach \i in {0,...,71}{%   a speech-like bar waveform: two bursts over a low floor
  \pgfmathsetmacro{\env}{min(1, 0.22+0.78*exp(-((\i-22)/8.5)^2)+0.9*exp(-((\i-50)/9.5)^2))}
  \pgfmathsetmacro{\h}{0.215*\env*(0.30+0.70*abs(sin(\i*103)*cos(\i*41)))}
  \draw[anuInk, line width=0.55pt, line cap=round] ({\cx-2.34+\i*0.066},{0.28-\h}) -- ++(0,{2*\h});
}
\node[lbl, anchor=north] at (\cx,-0.03) {16\,kHz audio $\cdot$ any length};
\node[box] (win) at (\cx,1.35) {16\,s windows $\cdot$ 1.6\,s hop};
\draw[arr] (wav) -- (win);
% ---------------- segmentation network: a wide frame with the encoder under the head ----------------
\def\fb{2.14}\def\bx{0.30}
\pgfmathsetmacro{\stop}{\fb+4.99}
\draw[anuBlue, line width=0.7pt, rounded corners=3pt, fill=white]
  ({\cx-\W/2-0.05},\fb) rectangle ({\cx+\W/2+0.05},{\stop+0.76});
% Actual depth: 24 pruned WavLM stages, then mixing/projection and four Conformer blocks.
\foreach \i in {1,...,24}{
  \pgfmathsetmacro{\y}{\fb+0.11+(\i-1)*0.125}
  \fill[anuStack, rounded corners=1pt] ({\cx-\W/2+0.12},\y) rectangle ({\cx+\W/2-0.12},{\y+0.10});
}
\node[box, minimum width=4.96cm, minimum height=0.55cm, font=\sffamily\fontsize{7}{8}\selectfont] (mix) at (\cx,{\fb+3.54})
  {Learned mix of 25 WavLM outputs\\Linear 1024$\to$256 $\cdot$ LayerNorm};
\draw[arr] (\cx,{\fb+3.085}) -- (mix.south);
\foreach \i in {0,...,3}{
  \pgfmathsetmacro{\y}{\fb+3.96+\i*0.27}
  \fill[anuBlue, rounded corners=1.5pt] ({\cx-\W/2+0.12},\y) rectangle ({\cx+\W/2-0.12},{\y+0.22});
}
\draw[arr] (mix.north) -- (\cx,{\fb+3.96});
\node[font=\sffamily\scriptsize, text=anuInk, anchor=north] at (\cx,{\stop+0.72}) {Segmentation network};
\node[lbl, anchor=north] at (\cx,{\stop+0.47}) {all parameters jointly fine-tuned};
\node[lbl, anchor=east, align=right] at ({\bx-0.06},{\fb+1.60}) {WavLM-Large\\24 pruned blocks\\80\,\% sparsity};
\node[lbl, text=anuBlue, anchor=east, align=right] at ({\bx-0.06},{\fb+4.475}) {Conformer head\\4 blocks, d = 256};
\draw[arr] (win.north) -- (\cx,\fb);
% ---------------- parent stages (bottom up) ----------------
\pgfmathsetmacro{\ya}{\stop+0.76+0.22+0.23}
\node[hero] (ps) at (\cx,\ya) {Powerset decisions $\cdot$ 16 classes};
\node[box, above=0.22cm of ps] (emb) {Embeddings of overlap-free frames};
\node[box, above=0.22cm of emb] (vbx) {VBx clustering of local speakers};
\node[box, above=0.22cm of vbx] (out) {Who spoke when};
\node[lbl, anchor=east, align=right] at ($(ps.west)+(-0.46,0)$) {4 local speakers\\20\,ms frames};
\node[lbl, anchor=west, align=left] at ($(emb.east)+(0.14,0)$) {ResNet34;\\6.6\,M};
\node[lbl, anchor=west, align=left] at ($(vbx.east)+(0.14,0)$) {parent PLDA;\\stock settings};
\draw[harr] (\cx,{\stop+0.76}) -- (ps.south);
\draw[arr] (ps) -- (emb); \draw[arr] (emb) -- (vbx); \draw[arr] (vbx) -- (out);
% ---------------- one block opened (every block is built the same way) ----------------
% main path up the centre of the modules with a residual add after each; each bypass runs on the left
\def\px{8.0}\def\pw{8.23}\def\pb{\fb}\pgfmathsetmacro{\pt}{\ya+3*0.46+2*0.22-0.23}
\draw[anuLine, line width=0.45pt, rounded corners=3pt, fill=anuChip] (\px,\pb) rectangle ({\px+\pw},\pt);
\node[font=\sffamily\scriptsize, text=anuInk, anchor=north west] at ({\px+0.18},{\pt-0.10}) {One Conformer block, opened};
\node[lbl, anchor=north west, align=left] at ({\px+0.18},{\pt-0.40}) {all four head blocks share this structure \citep{gulati2020conformer}};
\def\sx{11.15}\def\rx{8.80}\pgfmathsetmacro{\yb}{\pb+0.62}\pgfmathsetmacro{\dy}{(\pt-1.62-\yb)/4}
\foreach \k/\nm/\dt in {0/Feed-forward/half-step residual, 1/Multi-head self-attention/4 heads,
                        2/Convolution module/kernel 31, 3/Feed-forward/half-step residual, 4/LayerNorm/the block's output}{
  \node[sub] (m\k) at (\sx,{\yb+\k*\dy}) {\nm};
  \node[lbl, anchor=west] at ({\sx+2.12},{\yb+\k*\dy}) {\dt};
}
\foreach \k [evaluate=\k as \kn using int(\k+1)] in {0,...,3}{
  \node[circle, draw=anuMuted, line width=0.5pt, fill=white, inner sep=0pt, minimum size=0.23cm] (p\k)
    at (\sx,{\yb+\k*\dy+\dy/2}) {};
  \draw[anuMuted, line width=0.5pt] ($(p\k.center)+(-0.07,0)$) -- ++(0.14,0) ($(p\k.center)+(0,-0.07)$) -- ++(0,0.14);
  \draw[arr] (m\k.north) -- (p\k.south);
  \draw[arr] (p\k.north) -- (m\kn.south);
  \draw[arr] (\sx,{\yb+\k*\dy-0.40}) -| (\rx,{\yb+\k*\dy+\dy/2}) -- (p\k.west);   % residual bypass
}
\draw[arr] (\sx,{\pb+0.10}) -- (m0.south);
\node[lbl, anchor=west] at ({\sx+0.08},{\pb+0.13}) {from the preceding layer};
\draw[arr] (m4.north) -- ++(0,0.34);
\node[lbl, anchor=west] at ($(m4.north)+(0.08,0.24)$) {to the next layer};
% Open the second of four Conformer head blocks.
\draw[anuBlue, dash pattern=on 1.6pt off 1.2pt, line width=0.45pt]
  ({\cx+\W/2-0.12},{\fb+3.96+0.27+0.22}) -- (\px,\pt)
  ({\cx+\W/2-0.12},{\fb+3.96+0.27}) -- (\px,\pb);
% ---------------- three equal adaptation chips in the LID palette ----------------
\node[chip, anchor=north west, text width=5.31cm, minimum height=1.15cm] (c1) at (-0.62,-0.62) {\textbf{Sampling}\\50\,\% Indian calls, languages balanced\\25\,\% NOTSOFAR-1 $\cdot$ 25\,\% other meetings};
\node[chip, anchor=north west, text width=5.31cm, minimum height=1.15cm] (c2) at (5.057,-0.62) {\textbf{Labels}\\Per-speaker channels\\10\,ms cells marked known / unknown};
\node[chip, anchor=north west, text width=5.31cm, minimum height=1.15cm] (c3) at (10.733,-0.62) {\textbf{Optimisation}\\AdamW 2e-6 encoder, 2e-5 head\\3{,}000 updates $\times$ 96 crops $\cdot$ cosine $\cdot$ bf16};
% keep the chips at the same printed size as the LID figure's: pad the picture to the same natural width
\path ($(current bounding box.west)+(-1.075,0)$) ($(current bounding box.east)+(1.075,0)$);
\end{tikzpicture}}
  \caption{\sysname{}. The Indic-adapted segmentation network combines a pruned WavLM-Large encoder with four Conformer blocks (blue) and predicts 16 powerset classes every 20\,ms. The inset opens one Conformer block. Overlap-free speaker embeddings and VBx clustering link local speakers across windows; chips summarise the adaptation recipe.}
  \label{fig:system}
\end{figure*}
\else
\begin{figure}[t]
\centering
\resizebox{\columnwidth}{!}{%
\begin{tikzpicture}[
  font=\footnotesize,
  box/.style={draw=paperink, line width=0.5pt, rounded corners=2.5pt, minimum width=6.4cm, minimum height=0.68cm, align=center, fill=white},
  hero/.style={box, draw=ours, line width=0.9pt, fill=ours!7},
  arr/.style={-{Stealth[length=2.2mm]}, line width=0.6pt, draw=paperink},
  note/.style={font=\scriptsize, text=paperink!75, align=left, anchor=west}
]
\node[box] (wave) at (0,0) {\phantom{x}};
\begin{scope}
  \clip ($(wave.south west)+(0.15,0.08)$) rectangle ($(wave.north east)+(-0.15,-0.08)$);
  \draw[ours!80!black, line width=0.45pt, domain=-3.0:3.0, samples=420, smooth, variable=\x]
    plot ({\x}, {0.25*sin(deg(23*\x))*exp(-0.35*(\x+1.2)^2) + 0.18*sin(deg(31*\x+1))*exp(-0.5*(\x-1.4)^2) + 0.06*sin(deg(57*\x))});
\end{scope}
\node[note] at ($(wave.east)+(0.1,0)$) {audio, 16 kHz\\any length};
\node[hero, above=0.30cm of wave] (seg) {Segmentation: pruned WavLM-Large + Conformer\\powerset over 4 local speakers, 16 s windows};
\node[note] at ($(seg.east)+(0.1,0)$) {Indic\\adapted};
\node[box, above=0.30cm of seg] (emb) {Speaker embeddings of overlap free frames\\(ResNet34, per local speaker)};
\node[box, above=0.30cm of emb] (vbx) {VBx clustering of local speakers\\parent model's PLDA and default settings};
\node[box, above=0.30cm of vbx] (out) {Who spoke when};
\draw[arr] (wave) -- (seg); \draw[arr] (seg) -- (emb); \draw[arr] (emb) -- (vbx); \draw[arr] (vbx) -- (out);
\end{tikzpicture}}
\caption{\textbf{\sysname{} reads upward from the waveform.} Blue: the Indic adapted segmentation network, the contribution of this paper; the speaker embeddings and clustering are the parent model's.}
\label{fig:system}
\end{figure}
\fi

\sysname{} follows the EEND-VC design (Figure~\ref{fig:system}). A segmentation network processes 16\,s windows with a 1.6\,s hop and predicts, for every 20\,ms frame, one of 16 powerset classes over four local speakers. Overlap free frames of each local speaker are embedded with a ResNet34 speaker model \citep{he2016resnet,wang2023wespeaker}, local speakers are linked across windows by VBx clustering with the parent model's probabilistic linear discriminant analysis (PLDA) and default settings, and the frame level activities are aggregated into a recording level segmentation. Nothing after the network is tuned.

\paragraph{Starting point.}
\label{sec:adapt}
We start from DiariZen WavLM-Large s80 md-v2, whose segmentation network combines a WavLM-Large encoder pruned to 80\% sparsity with a Conformer head \citep{han2025diarizen,han2025pruning}, and adapt it to Indian conversational speech.

\paragraph{Data.}
The adaptation set contains 848 recordings (370.5 hours). Indian conversation comes from 302 two party telephone calls of SPRING-INX in eight languages (Assamese, Bengali, Gujarati, Hindi, Malayalam, Marathi, Punjabi, Tamil; 98.5 hours) \citep{nithya2023springinx}. Multi-party meetings come from NOTSOFAR-1 (71 recordings, 7.3 hours) \citep{vinnikov2024notsofar}, AMI (101, 59.9 hours) \citep{carletta2005ami}, AISHELL-4 (173, 97.2 hours) \citep{fu2021aishell4} and AliMeeting (201, 107.6 hours) \citep{yu2022m2met}. Each mini-batch draws 50\% of its crops from the Indian calls with languages balanced, 25\% from NOTSOFAR-1 and 25\% from the other meetings. The 302 calls are 90 calls admitted first and 212 of 346 later candidates: before admission, every candidate was screened against every other recording by acoustic fingerprinting \citep{chromaprint} with waveform corroboration, the 134 candidates that matched another recording were held out, and no audio in the adaptation set is shared with the benchmark.

\paragraph{Silver labels with explicit uncertainty.}
Each SPRING-INX call was recorded with one channel per speaker, which yields speaker labels without manual annotation, but channel evidence is imperfect: a quiet speaker, crosstalk or untranscribed speech can make a label wrong. Rather than trusting or discarding such labels, we mark every 10\,ms cell as known or unknown. A cell is unknown if a positive label lacks energy support on its own channel, is more than 15\,dB quieter than the other channel, contradicts a transcript interval, falls within 100\,ms of a label or transcript boundary, or disagrees with an independent pretrained diarizer applied to each isolated channel. Unknown frames contribute no loss; quiet positives are masked rather than relabeled as silence. Of the 98.5 hours of call audio, 51.9 hours remain supervised.

\paragraph{Training.}
We train all parameters jointly for 3,000 updates of 96 sixteen second crops each (288,000 crops) with AdamW ($\beta=(0.9, 0.98)$, weight decay 0.01), learning rates of $2\times10^{-6}$ for the encoder and $2\times10^{-5}$ for the head, 50 warmup updates, cosine decay, gradient clipping at 1.0 and bfloat16 arithmetic. Training takes about 91 minutes on six RTX A6000 GPUs.

\paragraph{Size.}
The network has 69.5\,M parameters and the speaker model 6.6\,M. Inference is deterministic: rerunning it reproduces its outputs byte for byte.

% ===================================================================================================
\ifaclpreprint
% Declare the wide result table in time for the following Results page.
\begin{table*}[t]
\centering
\footnotesize
\setlength{\tabcolsep}{1.75pt}
\begin{tabular}{@{}l r r r r r @{\hspace{7pt}} l r r r r@{}}
\toprule
\multicolumn{6}{c}{\textbf{Run under our protocol}} & \multicolumn{5}{c}{\textbf{Reported by \citet{mehendale2026indicdiarbench}}} \\
\cmidrule(r){1-6}\cmidrule(l){7-11}
\textbf{System} & \textbf{Pool.} & \textbf{DW} & \textbf{Miss} & \textbf{FA} & \textbf{Conf.} & \textbf{System} & \textbf{DW} & \textbf{Miss} & \textbf{FA} & \textbf{Conf.} \\
\midrule
\textbf{Ours (\sysname{})} & \textbf{17.11} & \textbf{16.78} & 8.91 & 3.55 & 4.65 & Sarvam AI$^{\dagger}$ & 16.0 & 6.3 & 3.9 & 5.9 \\
pyannoteAI Precision-2 & 17.58 & 17.18 & 9.70 & 3.06 & 4.82 & AWS Transcribe$^{\dagger}$ & 23.5 & 13.1 & 3.1 & 7.4 \\
DiariZen WavLM-Large s80 md-v2 & 18.31 & 17.99 & 11.07 & 3.03 & 4.22 & Deepgram Nova-3$^{\dagger}$ & 32.0 & 18.3 & 5.4 & 8.3 \\
NVIDIA Nemotron-3-Diarization & 18.71 & 18.38 & 13.07 & 2.63 & 3.01 & Azure STT$^{\dagger}$ & 34.8 & 24.4 & 1.7 & 8.7 \\
pyannote speaker-diarization-community-1 & 21.65 & 21.22 & 11.25 & 2.94 & 7.46 & ElevenLabs Scribe$^{\dagger}$ & 35.0 & 13.6 & 6.2 & 15.3 \\
NVIDIA Streaming Sortformer 4-spk v2.1 & 22.34 & 22.14 & 8.94 & 4.32 & 9.07 & GPT-4o Transcribe$^{\dagger}$ & 36.2 & 17.7 & 3.8 & 14.7 \\
DiCoW v3.3 & 22.76 & 22.62 & 13.58 & 5.44 & 3.73 & AssemblyAI Universal-2$^{\dagger}$ & 40.5 & 25.5 & 5.5 & 9.6 \\
3D-Speaker (CAM++ pipeline) & 24.39 & 23.67 & 13.01 & 4.30 & 7.08 & Gemini 3 Pro$^{\dagger}$ & 74.0 & 41.7 & 5.8 & 26.5 \\
MOSS-Transcribe-Diarize & 34.52 & 33.76 & 25.11 & 1.74 & 7.67 &  & & & &  \\
NVIDIA Sortformer 4-spk v1$^{b}$ & 39.66 & 39.41 & 30.68 & 2.30 & 6.68 &  & & & &  \\
 
\bottomrule
\end{tabular}
\caption{\textbf{Full IndicDiarBench results} (1,164 recordings; DER in \%, zero collar, overlap scored). Left: run under our protocol, pooled (Pool.) and duration weighted (DW); ours in bold. Right: joint transcription and diarization services ($^{\dagger}$), as reported and named by the benchmark's authors, who do not give model versions for all of them. Every system was evaluated zero shot; 4-spk: 4-speaker; Miss: missed speech; FA: false alarm; Conf.: confusion. $^{b}$Inputs beyond the model's maximum length are scored as missed.}
\label{tab:main}
\end{table*}
\fi

\ifaclpreprint
\subsection{Inference and reproducibility}
\label{app:details}

\paragraph{Inference.} Windows of 16\,s with a 1.6\,s hop; powerset decisions by argmax followed by an 11 frame median filter; embeddings from overlap free frames when a local speaker has more than two such frames; agglomerative initialization (threshold 0.6) followed by VBx with the parent model's PLDA and parameters ($F_a=0.07$, $F_b=0.8$, 20 iterations). Inference uses FP32 arithmetic with TF32 disabled, deterministic kernels and a fixed seed reset for every file.

\paragraph{Data, models and software.} IndicDiarBench is licensed CC BY 4.0. NOTSOFAR-1, AMI and VoxConverse are licensed CC BY 4.0, AISHELL-4 and AliMeeting CC BY-SA 4.0, and MagicData-RAMC CC BY-NC-ND 4.0 (used for evaluation only); SPRING-INX is released for research by its creators \citep{nithya2023springinx}. The parent segmentation weights are licensed CC BY-NC 4.0, the ResNet34 speaker model and the PLDA CC BY 4.0, and pyannote speaker-diarization-community-1 CC BY 4.0; the other competitor systems were used under their own terms. We use PyTorch 2.1.1, pyannote.audio 3.1.1 in the DiariZen code base (commit 844f555), scikit-learn 1.7.2 and NumPy 1.26.4.

\paragraph{Reproducibility.} The supplementary material contains the score files and the scripts that regenerate every table and figure of this paper from them.
\fi

\section{Results}
\label{sec:results}

\ifaclpreprint
\else
\begin{table*}[t]
\centering
\footnotesize
\setlength{\tabcolsep}{1.75pt}
\begin{tabular}{@{}l r r r r r @{\hspace{7pt}} l r r r r@{}}
\toprule
\multicolumn{6}{c}{\textbf{Run under our protocol}} & \multicolumn{5}{c}{\textbf{Reported by \citet{mehendale2026indicdiarbench}}} \\
\cmidrule(r){1-6}\cmidrule(l){7-11}
\textbf{System} & \textbf{Pool.} & \textbf{DW} & \textbf{Miss} & \textbf{FA} & \textbf{Conf.} & \textbf{System} & \textbf{DW} & \textbf{Miss} & \textbf{FA} & \textbf{Conf.} \\
\midrule
\textbf{Ours (\sysname{})} & \textbf{17.11} & \textbf{16.78} & 8.91 & 3.55 & 4.65 & Sarvam AI$^{\dagger}$ & 16.0 & 6.3 & 3.9 & 5.9 \\
pyannoteAI Precision-2 & 17.58 & 17.18 & 9.70 & 3.06 & 4.82 & AWS Transcribe$^{\dagger}$ & 23.5 & 13.1 & 3.1 & 7.4 \\
DiariZen WavLM-Large s80 md-v2 & 18.31 & 17.99 & 11.07 & 3.03 & 4.22 & Deepgram Nova-3$^{\dagger}$ & 32.0 & 18.3 & 5.4 & 8.3 \\
NVIDIA Nemotron-3-Diarization & 18.71 & 18.38 & 13.07 & 2.63 & 3.01 & Azure STT$^{\dagger}$ & 34.8 & 24.4 & 1.7 & 8.7 \\
pyannote speaker-diarization-community-1 & 21.65 & 21.22 & 11.25 & 2.94 & 7.46 & ElevenLabs Scribe$^{\dagger}$ & 35.0 & 13.6 & 6.2 & 15.3 \\
NVIDIA Streaming Sortformer 4-spk v2.1 & 22.34 & 22.14 & 8.94 & 4.32 & 9.07 & GPT-4o Transcribe$^{\dagger}$ & 36.2 & 17.7 & 3.8 & 14.7 \\
DiCoW v3.3 & 22.76 & 22.62 & 13.58 & 5.44 & 3.73 & AssemblyAI Universal-2$^{\dagger}$ & 40.5 & 25.5 & 5.5 & 9.6 \\
3D-Speaker (CAM++ pipeline) & 24.39 & 23.67 & 13.01 & 4.30 & 7.08 & Gemini 3 Pro$^{\dagger}$ & 74.0 & 41.7 & 5.8 & 26.5 \\
MOSS-Transcribe-Diarize & 34.52 & 33.76 & 25.11 & 1.74 & 7.67 &  & & & &  \\
NVIDIA Sortformer 4-spk v1$^{b}$ & 39.66 & 39.41 & 30.68 & 2.30 & 6.68 &  & & & &  \\
 
\bottomrule
\end{tabular}
\caption{\textbf{Full IndicDiarBench results} (1,164 recordings; DER in \%, zero collar, overlap scored). Left: run under our protocol, pooled (Pool.) and duration weighted (DW); ours in bold. Right: joint transcription and diarization services ($^{\dagger}$), as reported and named by the benchmark's authors, who do not give model versions for all of them. Every system was evaluated zero shot; 4-spk: 4-speaker; Miss: missed speech; FA: false alarm; Conf.: confusion. $^{b}$Inputs beyond the model's maximum length are scored as missed.}
\label{tab:main}
\end{table*}
\fi

\ifaclpreprint
\subsection{Overall benchmark performance}
\else
\paragraph{Results on IndicDiarBench.}
\fi
Table~\ref{tab:main} and Figure~\ref{fig:leaderboard} show that \sysname{} reaches 17.11\% pooled and 16.78\% duration weighted DER on all 1,164 recordings (pooled 95\% CI: 15.80 to 18.45), the lowest of all systems we ran. The paired bootstrap differences all exclude zero: $-$0.47\pp{} (95\% CI [$-$0.84, $-$0.09]) against pyannoteAI Precision-2, $-$1.21 ([$-$1.40, $-$1.01]) against DiariZen md-v2 and $-$1.60 ([$-$2.18, $-$1.06]) against NVIDIA Nemotron-3-Diarization. The only lower published figure is the 16.0\% self-reported by Sarvam AI for its joint transcription and diarization service, 0.78\pp{} below ours in duration weighted terms. Every system is a single trained model, so the intervals reflect the sampling of recordings, not training seeds.

\ifaclpreprint
\subsection{Effect of Indic adaptation}
\else
\paragraph{From the parent model to \sysname{}.}
\fi
Indic adaptation lowers DER from 18.31\% to 17.11\% (duration weighted: 17.99\% to 16.78\%; Table~\ref{tab:ablation}). It recovers missed speech (11.07\% to 8.91\%) at a cost of 0.52\pp{} more false alarm and 0.43\pp{} more confusion. The third row is the zero shot convention adapter of Section~\ref{sec:transfer}.

\ifaclpreprint
\begin{table}[!htb]
\else
\begin{table}[t]
\fi
\centering
\footnotesize
\setlength{\tabcolsep}{2.2pt}
\begin{tabular}{@{}>{\raggedright\arraybackslash}p{3.3cm} r r r r r@{}}
\toprule
\textbf{Configuration} & \textbf{Pool.} & \textbf{DW} & \textbf{Miss} & \textbf{FA} & \textbf{Conf.} \\
\midrule
DiariZen WavLM-Large s80 md-v2 (parent) & 18.31 & 17.99 & 11.07 & 3.03 & 4.22 \\
+ Indic adaptation (\sysname{}) & \textbf{17.11} & \textbf{16.78} & 8.91 & 3.55 & 4.65 \\
+ convention learned from telephone transcripts$^{z}$ & 22.74 & 22.43 & 2.89 & 15.29 & 4.56 \\
 
\bottomrule
\end{tabular}
\caption{\textbf{Indic adaptation helps; a convention learned from telephone transcripts does not transfer.} DER (\%) on all 1,164 recordings, each row evaluated zero shot. $^{z}$Fitted on 302 SPRING-INX calls and selected on 35 others under a fixed protocol; scored once (\technicalref{app:zs}).}
\label{tab:ablation}
\end{table}

\ifaclpreprint
\subsection{Languages, conditions and speaker counts}
\label{app:breakdown}
\else
\paragraph{Languages, conditions and speaker counts.}
\fi
\sysname{} leads the three strongest other systems in 14 of the 22 languages (Table~\ref{tab:languages}). Its largest margins are in Manipuri ($-$1.92\pp{}), Sanskrit ($-$1.88), Urdu ($-$1.51) and Santali ($-$1.33), none of which is in our adaptation data; it trails the best other system most in Maithili ($+$1.98), Sindhi ($+$1.80) and Bodo ($+$1.56). Of these four systems, it is the most accurate in near field meetings (19.05\% against 19.93\% for the best other system) and far field meetings (18.64\% against 19.48\%), but not on in-the-wild video, where pyannoteAI Precision-2 leads (10.78\% against 11.51\%). It leads at every number of speakers from two to eight except four (\ifaclpreprint Figure~\ref{fig:structure}, Table~\ref{tab:breakdown}\else Appendix~\ref{app:breakdown}, Figure~\ref{fig:structure}\fi).

\ifaclpreprint

\begin{figure*}[t]
\centering
\includegraphics[width=\textwidth]{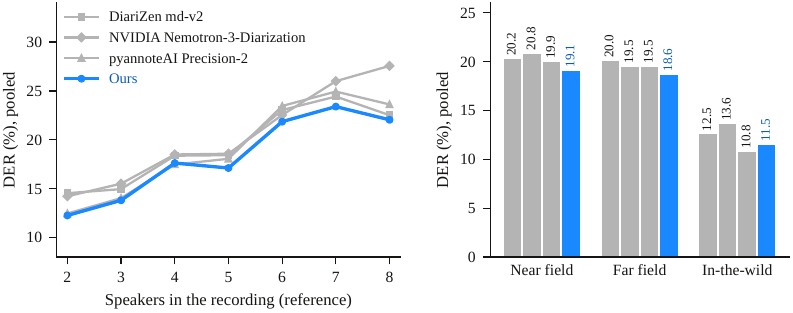}
\caption{\textbf{\sysname{} leads in near field and far field meetings and at six of seven speaker counts.} Pooled DER on all 1,164 recordings by the number of reference speakers (left; 2 to 8 speakers cover 1,157 recordings) and by recording condition (right; bars left to right: DiariZen md-v2, NVIDIA Nemotron-3-Diarization, pyannoteAI Precision-2, \sysname{}).}
\label{fig:structure}
\end{figure*}

\begin{table}[!htb]
\centering
\small
\setlength{\tabcolsep}{3.5pt}
\begin{tabular}{@{}l r r r r r@{}}
\toprule
\textbf{Subset} & \textbf{Rec.} & \textbf{Ours} & \textbf{P2} & \textbf{DZ} & \textbf{NM} \\
\midrule
Near field & 781 & \textbf{19.05} & 19.93 & 20.25 & 20.83 \\
Far field & 173 & \textbf{18.64} & 19.48 & 20.03 & 19.48 \\
In-the-wild & 210 & 11.51 & \textbf{10.78} & 12.55 & 13.58 \\
2 speakers & 294 & \textbf{12.26} & 12.46 & 14.54 & 14.22 \\
3 speakers & 227 & \textbf{13.81} & 14.03 & 14.96 & 15.51 \\
4 speakers & 234 & 17.62 & \textbf{17.48} & 18.38 & 18.50 \\
5 speakers & 148 & \textbf{17.12} & 18.06 & 18.43 & 18.58 \\
6 speakers & 162 & \textbf{21.86} & 23.46 & 23.05 & 22.54 \\
7 speakers & 66 & \textbf{23.40} & 24.94 & 24.43 & 26.00 \\
8 speakers & 26 & \textbf{22.05} & 23.62 & 22.52 & 27.56 \\
 
\bottomrule
\end{tabular}
\caption{Pooled DER (\%) by recording condition (all 1,164 recordings) and by number of reference speakers (1,157 recordings with 2--8 speakers). Rec.: recordings; P2: pyannoteAI Precision-2; DZ: DiariZen md-v2; NM: NVIDIA Nemotron-3-Diarization. Best in bold.}
\label{tab:breakdown}
\end{table}

Table~\ref{tab:breakdown} gives the recording counts behind these comparisons. The speaker-count analysis covers 1,157 recordings with two to eight speakers; the seven recordings with nine speakers remain in the full benchmark scores. The groups differ in both size and recording content: there are 294 two-speaker recordings but only 26 eight-speaker recordings. The trend therefore describes performance on the observed groups, rather than a controlled effect of adding speakers to the same conversation.
\fi

\ifaclpreprint
\subsection{Transfer to international benchmarks}
\else
\paragraph{Beyond IndicDiarBench.}
\fi
\label{sec:beyond}
A model adapted to one population can lose ground elsewhere. On six international test sets (Table~\ref{tab:public}\ifaclpreprint\else, Appendix~\ref{app:breakdown}\fi), \sysname{} matches its parent: it is better on three sets and worse on three, with a mean DER of 12.03\% against 12.19\% for DiariZen md-v2. Indic adaptation thus costs no accuracy on these sets on average.

% ===================================================================================================
\ifaclpreprint

\begin{table}[!htb]
\centering
\small
\setlength{\tabcolsep}{3pt}
\begin{tabular}{@{}l r r r@{}}
\toprule
& & \textbf{DiariZen} & \\
\textbf{Test set} & \textbf{Files} & \textbf{md-v2} & \textbf{Ours} \\
\midrule
VoxConverse v0.3 test & 232 & \textbf{9.17} & 9.32 \\
NOTSOFAR-1 eval$^{a}$ & 129 & 18.26 & \textbf{17.41} \\
MagicData-RAMC test & 43 & \textbf{10.97} & 11.13 \\
AMI test (SDM) & 16 & \textbf{13.93} & 14.00 \\
AISHELL-4 test & 20 & 10.06 & \textbf{9.85} \\
AliMeeting test (far) & 20 & 10.74 & \textbf{10.44} \\
\midrule
Mean & & 12.19 & \textbf{12.03} \\
 
\bottomrule
\end{tabular}
\caption{\textbf{The adapted network matches its parent beyond India.} DER (\%), zero collar, overlap scored, both with the parent model's clustering \citep{chung2020voxconverse,vinnikov2024notsofar,yang2022ramc,carletta2005ami,fu2021aishell4,yu2022m2met}. The better of the two is in bold. SDM: single distant microphone. $^{a}$A single channel device per meeting.}
\label{tab:public}
\end{table}

The mean in Table~\ref{tab:public} gives each test set equal weight; it is not pooled across all recordings or hours. The largest improvement over the parent is 0.85\pp{} on NOTSOFAR-1, and the largest regression is 0.16\pp{} on MagicData-RAMC. Both systems use the parent model's clustering, so these comparisons evaluate the effect of changing the segmentation network within the same inference pipeline.
\FloatBarrier
\fi

\section{Discussion}
\label{sec:discussion}

\paragraph{Convention is part of the task.}
The convention adapter failed for a reason that supports Section~\ref{sec:convention}: two corpora transcribed into speaker attributed utterances still disagree about where an utterance ends. Matching a benchmark's convention therefore needs references made the way the benchmark's are, or a convention documented precisely enough to be reproduced.

\paragraph{Linguistic diversity and transfer.}
\sysname{}'s Indian adaptation data covers eight languages of two party telephone conversation, yet ten of the fourteen languages in which it leads are languages it never saw in adaptation, including Manipuri, Sanskrit, Urdu and Santali. This suggests that the adaptation mostly teaches the acoustic and conversational conditions of Indian speech, which transfer across the families of the linguistic area, rather than properties of particular languages. Difficulty remains where conversation is hardest: near field debates with many participants and over 20\% overlapped speech, as in Nepali, Telugu, Maithili and Dogri.

% ===================================================================================================
\section{Conclusion}

We presented the first systematic study of dedicated speaker diarization systems across India's 22 scheduled languages, showed that the benchmark's utterance level convention is a hidden variable worth 1.9 to 4.1\pp{} of DER and that a convention learned from other transcriptions does not transfer to it, and introduced \sysname{}, which, evaluated zero shot, is the most accurate of all systems we ran (16.78\% duration weighted DER) and matches its parent across six international benchmarks.
% Release statement, to be confirmed by the authors before submission:
% We release the system, its evaluation protocol and the scripts that produce every number in this paper.

% ===================================================================================================
\section*{Limitations}

\sysname{} does not model the benchmark's segmentation convention: matching it would lower its DER by about 1.9\pp{} (Table~\ref{tab:convention}), but our attempt to learn it from other transcriptions failed (\S\ref{sec:transfer}), and we did not fit it on the benchmark itself. Its duration weighted DER remains 0.78\pp{} above the 16.0\% self-reported by Sarvam AI, whose system we did not run. The segmentation network models at most four simultaneous speakers within a 16\,s window, although the number of speakers per recording is unbounded. The Indian adaptation data comprises two party telephone calls in eight languages; multi-party Indian meetings for training would likely help further. Our evaluation uses one benchmark of 108 hours; the international results provide a second view but not an Indian one. The network inherits the non-commercial license (CC BY-NC 4.0) of its parent model's weights.

\section*{Ethical Considerations}

All training and evaluation data are publicly released research corpora, used under their terms of release; no new recordings of people were made. Diarization identifies when different people speak but not who they are; it can nevertheless be combined with other tools to profile speakers, and the model's weights are licensed for non-commercial research use only. Performance varies across languages and conditions (Table~\ref{tab:languages}); deployments should evaluate on their own population.

% ---------------------------------------------------------------------------------------------------

% Keep all preprint method/results floats before the reference list.
\ifaclpreprint\FloatBarrier\fi
\bibliography{references}

% ===================================================================================================
\ifaclpreprint
\else
\appendix

\begin{figure*}[t]
\centering
\includegraphics[width=\textwidth]{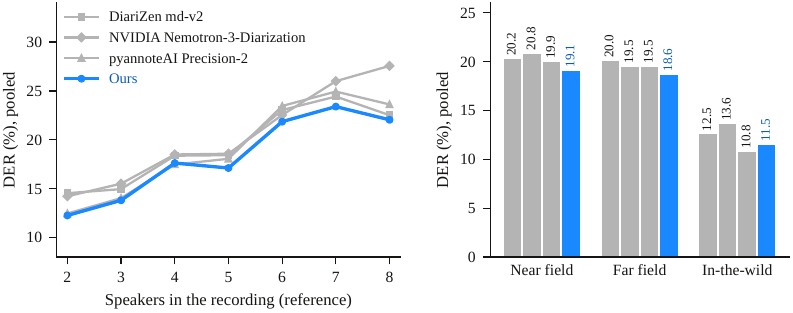}
\caption{\textbf{\sysname{} leads in near field and far field meetings and at six of seven speaker counts.} Pooled DER on all 1,164 recordings by the number of reference speakers (left; 2 to 8 speakers cover 1,157 recordings) and by recording condition (right; bars left to right: DiariZen md-v2, NVIDIA Nemotron-3-Diarization, pyannoteAI Precision-2, \sysname{}).}
\label{fig:structure}
\end{figure*}

\section{Implementation Details}
\label{app:details}

\paragraph{Inference.} Windows of 16\,s with a 1.6\,s hop; powerset decisions by argmax followed by an 11 frame median filter; embeddings from overlap free frames when a local speaker has more than two such frames; agglomerative initialization (threshold 0.6) followed by VBx with the parent model's PLDA and parameters ($F_a=0.07$, $F_b=0.8$, 20 iterations). Inference uses FP32 arithmetic with TF32 disabled, deterministic kernels and a fixed seed reset for every file.

\paragraph{Scoring.} DER with no collar and overlap scored, NIST md-eval semantics; pooled DER divides total error time by total reference speaker time; duration weighted DER averages per recording DER with weights proportional to recording duration. Bootstrap intervals resample the 590 source recordings 5,000 times (seed 3407).

\paragraph{Zero shot convention adapter.}
\label{app:zs}
A protocol fixing every candidate and decision rule was committed before any measurement, and a filesystem guard refused every IndicDiarBench file to the selection code. For this experiment only, clustering was chosen on 146 recordings of NOTSOFAR-1, VoxConverse, SPRING-INX, AMI, AISHELL-4 and AliMeeting from 21 VBx configurations, with and without a cannot-link repair that splits clusters whose strongly supported local speakers share a window; the repair at threshold 0.65 and $F_a=0.06$ won (mean pooled DER 12.26\% against 12.51\% for the default). References for the convention were the human transcript spans of each SPRING-INX call's two channels. The bridging gap was chosen on 35 held out calls from $g\in\{0, 0.25, \dots, 1.5\}$\,s (best: 1.5\,s). The decoder, gradient boosted regression trees \citep{ke2017lightgbm,pedregosa2011sklearn} with 400 iterations, learning rate 0.05, 31 leaves and at least 200 samples per leaf, predicts for every same speaker gap and every segment shorter than 3\,s the change in error time that bridging the gap or dropping the segment would cause, from the local geometry, the recording and the network's own evidence. Fitted on the 302 training calls, it beat blanket bridging on the 35 held out calls (duration weighted DER 30.02\% against 32.11\%, paired 95\% CI entirely below zero) and was adopted. The frozen system was scored once on IndicDiarBench: 22.74\% pooled and 22.43\% duration weighted DER.

\paragraph{Data, models and software.} IndicDiarBench is licensed CC BY 4.0. NOTSOFAR-1, AMI and VoxConverse are licensed CC BY 4.0, AISHELL-4 and AliMeeting CC BY-SA 4.0, and MagicData-RAMC CC BY-NC-ND 4.0 (used for evaluation only); SPRING-INX is released for research by its creators \citep{nithya2023springinx}. The parent segmentation weights are licensed CC BY-NC 4.0, the ResNet34 speaker model and the PLDA CC BY 4.0, and pyannote speaker-diarization-community-1 CC BY 4.0; the other competitor systems were used under their own terms. We use PyTorch 2.1.1, pyannote.audio 3.1.1 in the DiariZen code base (commit 844f555), scikit-learn 1.7.2 and NumPy 1.26.4.

\paragraph{Reproducibility.} The supplementary material contains the score files and the scripts that regenerate every table and figure of this paper from them.

\section{Additional Results}
\label{app:breakdown}

Figure~\ref{fig:structure} and Table~\ref{tab:breakdown} break the full benchmark results down by recording condition and number of speakers, and Table~\ref{tab:public} reports the international test sets.

\begin{table}[!htb]
\centering
\small
\setlength{\tabcolsep}{3.5pt}
\begin{tabular}{@{}l r r r r r@{}}
\toprule
\textbf{Subset} & \textbf{Rec.} & \textbf{Ours} & \textbf{P2} & \textbf{DZ} & \textbf{NM} \\
\midrule
Near field & 781 & \textbf{19.05} & 19.93 & 20.25 & 20.83 \\
Far field & 173 & \textbf{18.64} & 19.48 & 20.03 & 19.48 \\
In-the-wild & 210 & 11.51 & \textbf{10.78} & 12.55 & 13.58 \\
2 speakers & 294 & \textbf{12.26} & 12.46 & 14.54 & 14.22 \\
3 speakers & 227 & \textbf{13.81} & 14.03 & 14.96 & 15.51 \\
4 speakers & 234 & 17.62 & \textbf{17.48} & 18.38 & 18.50 \\
5 speakers & 148 & \textbf{17.12} & 18.06 & 18.43 & 18.58 \\
6 speakers & 162 & \textbf{21.86} & 23.46 & 23.05 & 22.54 \\
7 speakers & 66 & \textbf{23.40} & 24.94 & 24.43 & 26.00 \\
8 speakers & 26 & \textbf{22.05} & 23.62 & 22.52 & 27.56 \\
 
\bottomrule
\end{tabular}
\caption{Pooled DER (\%) by recording condition and by number of reference speakers on all 1,164 recordings. Rec.: recordings; P2: pyannoteAI Precision-2; DZ: DiariZen md-v2; NM: NVIDIA Nemotron-3-Diarization. Best in bold.}
\label{tab:breakdown}
\end{table}

\begin{table}[!htb]
\centering
\small
\setlength{\tabcolsep}{3pt}
\begin{tabular}{@{}l r r r@{}}
\toprule
& & \textbf{DiariZen} & \\
\textbf{Test set} & \textbf{Files} & \textbf{md-v2} & \textbf{Ours} \\
\midrule
VoxConverse v0.3 test & 232 & \textbf{9.17} & 9.32 \\
NOTSOFAR-1 eval$^{a}$ & 129 & 18.26 & \textbf{17.41} \\
MagicData-RAMC test & 43 & \textbf{10.97} & 11.13 \\
AMI test (SDM) & 16 & \textbf{13.93} & 14.00 \\
AISHELL-4 test & 20 & 10.06 & \textbf{9.85} \\
AliMeeting test (far) & 20 & 10.74 & \textbf{10.44} \\
\midrule
Mean & & 12.19 & \textbf{12.03} \\
 
\bottomrule
\end{tabular}
\caption{\textbf{The adapted network matches its parent beyond India.} DER (\%), zero collar, overlap scored, both with the parent model's clustering \citep{chung2020voxconverse,vinnikov2024notsofar,yang2022ramc,carletta2005ami,fu2021aishell4,yu2022m2met}. The better of the two is in bold. SDM: single distant microphone. $^{a}$A single channel device per meeting.}
\label{tab:public}
\end{table}

\fi

\end{document}